\documentclass[10pt,twocolumn,letterpaper]{article}

\usepackage{wacv}
\usepackage{times}
\usepackage{epsfig}
\usepackage{graphicx}
\usepackage{amsmath}
\usepackage{amssymb}

\usepackage{tgcursor}
\usepackage{multirow}
\usepackage{makecell}
\usepackage{booktabs}
\usepackage[hyphens]{url}
\newcolumntype{L}[1]{>{\raggedright\let\newline\\\arraybackslash\hspace{0pt}}m{#1}}
\newcolumntype{C}[1]{>{\centering\let\newline\\\arraybackslash\hspace{0pt}}m{#1}}
\newcolumntype{R}[1]{>{\raggedleft\let\newline\\\arraybackslash\hspace{0pt}}m{#1}}
\newcommand{\RED}[1]{\textcolor{black}{#1}}

\wacvfinalcopy % *** Uncomment this line for the final submission
\def\wacvPaperID{172} % *** Enter the wacv Paper ID here

\ifwacvfinal\fi
\begin{document}

%%%%%%%%% TITLE
\title{Beyond Pixels: Image Provenance Analysis Leveraging Metadata}

% Authors at the same institution
%\author{First Author \hspace{2cm} Second Author \\
% Authors at different institutions
% \author{First Author \\
% \and
% Second Author \\

\author{Aparna Bharati\(^1\), Daniel Moreira\(^1\), Joel Brogan\(^1\), Patricia Hale\(^1\),\\ Kevin W. Bowyer\(^1\), Patrick J. Flynn\(^1\), Anderson Rocha\(^2\), Walter J. Scheirer\(^1\)\\
\(^1\) University of Notre Dame, IN, USA \hspace{0.5cm}
\(^2\)University of Campinas, SP, Brazil\\
%{\tt\small \(^1\)\{abharati, dhenriq1, jbrogan4, phale1, kwb, flynn, wscheire\}@nd.edu\\ 
%\tt\small \(^2\)\{anderson.rocha\}@ic.unicamp.br}
}

\maketitle
\ifwacvfinal\thispagestyle{empty}\fi

%%%%%%%%% ABSTRACT
\begin{abstract}
   Creative works, whether paintings or memes, follow unique journeys that result in their final form. Understanding these journeys, a process known as ``provenance analysis," provides rich insights into the use, motivation, and authenticity underlying any given work. The application of this type of study to the expanse of unregulated content on the Internet is what we consider in this paper. Provenance analysis provides a snapshot of the chronology and validity of content as it is uploaded, re-uploaded, and modified over time. Although still in its infancy, automated provenance analysis for online multimedia is already being applied to different types of content.  Most current works seek to build provenance graphs based on the shared content between images or videos. This can be a computationally expensive task, especially when considering the vast influx of content that the Internet sees every day. Utilizing non-content-based information, such as timestamps, geotags, and camera IDs can help provide important insights into the path a particular image or video has traveled during its time on the Internet without large computational overhead. This paper\footnote{This material is based on research sponsored by DARPA and Air Force Research Laboratory (AFRL) under agreement number FA8750-16-2-0173. Hardware support was generously provided by the NVIDIA Corporation. We also thank the financial support of FAPESP (Grant 2017/12646-3, D\'ej\`aVu Project), CAPES (DeepEyes Grant) and CNPq (Grant 304472/2015-8).} tests the scope and applicability of metadata-based inferences for provenance graph construction in two different scenarios: digital image forensics and cultural analytics.
   %\RED{Results suggest that, if reliable metadata are available, they help to improve the quality of the generated provance graphs}

\end{abstract}
\vspace{-10pt}
%%%%%%%%% BODY TEXT
\section{Introduction}
\label{sec:intro}
Understanding the story behind a visual object is an activity of broad interest. Whether it is determining the palette used to make a painting, the style of a sculptor, or the authenticity of an artwork, deriving the origin and composition of the object at hand has been a difficult but important task for many examiners. Subtle clues derived from the nature of works of art have long been used to provide answers to \textit{provenance} related questions~\cite{bbcprogram}. Off-white colors found in the painting \emph{Darby and Joan} by Laurence Stephen Lowry brought into question its authenticity \cite{fakeorfortune_2015}. Lead content in the paint of \emph{Danseuse Bleue et Contrebasses} and careful scrutiny of the painter's signature allowed experts to rightly restore the validity of Edgar Degas's most famous work \cite{fakeorfortune_2012}. %DNA analysis of a stray hair was used to strengthen the claims that a portrait of Freud had been painted by Freud himself \cite{fakeorfortune_2016}. 
Provenance analysis of this sort has helped historians, cultural analysts and art enthusiasts to analyze the origin, content and growth of works such as these.  Although the techniques used to perform provenance analysis have evolved over time~\cite{douglas2010origins}, it is, in general, still an unsolved problem~\cite{provimpnews}. In the domain of art history, it is one of the most active and important areas of research~\cite{metprovproject} as there are still complicated cases where provenance has yet to be established (\textit{e.g.}, the painting \textit{Bords de la Seine \`{a} Argenteuil}~\cite{monetbords}) and new avenues for the interpretation of relationships between artworks.

%In addition to that, Whether it is based on a forensic evidence such as DNA in the hair stuck to an oil painting (``Freud") or notes from the seller, provenance has used a Though the techniques over

\begin{figure}[t]
\begin{center}
   \includegraphics[width=\linewidth]{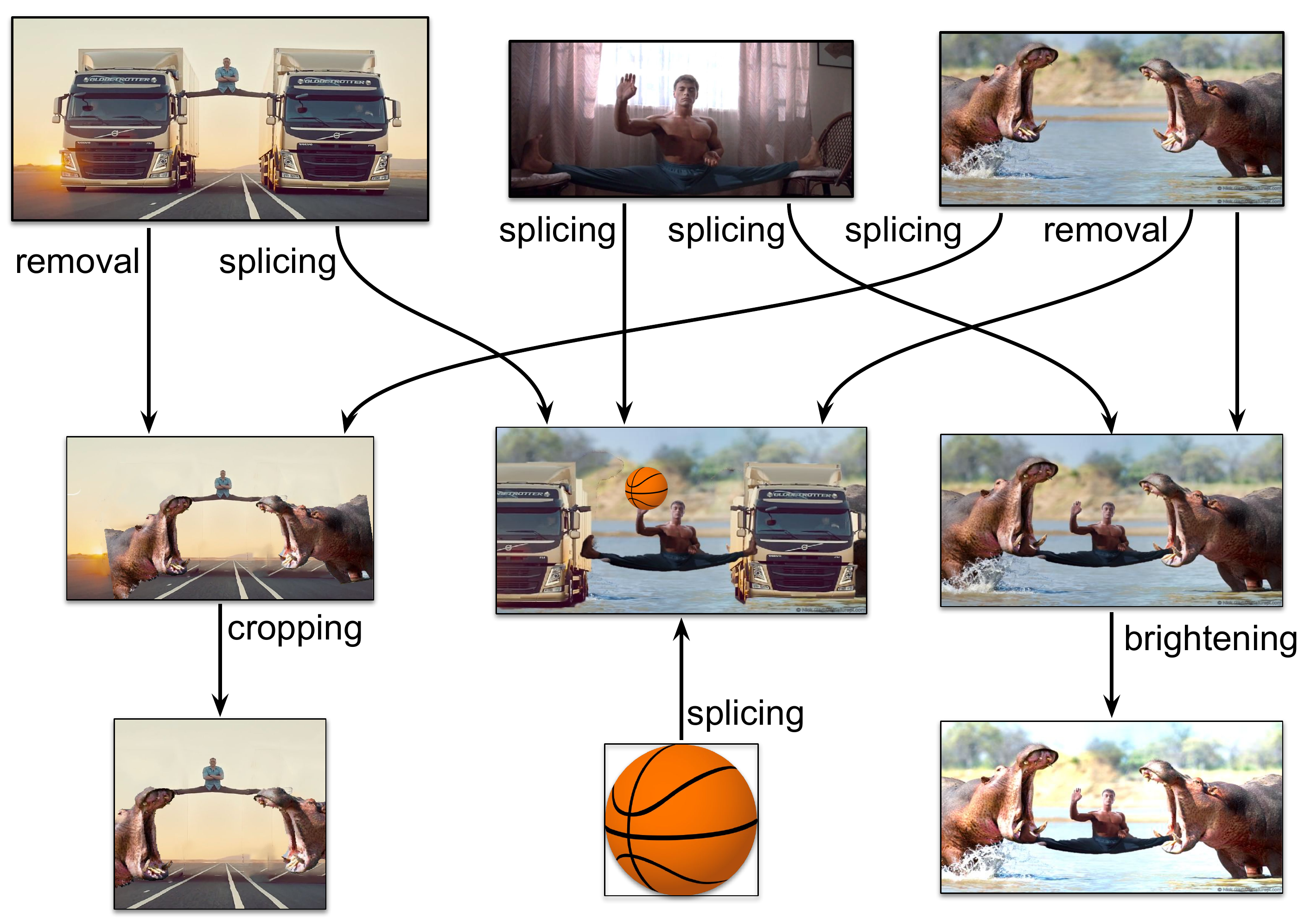} %2_bordered.pdf}
\end{center}
   \caption{Example of an Image Provenance Graph (IPG) showing some common operations performed on images and how they are manifested in the case of provenance. The examples in this case are meme-style images %from the /r/photoshopbattles community
   similar to the ones from the \emph{photoshopbattles} community on the Reddit social media site~\cite{reddit2017photoshopbattles}. The transformations can be as simple as increasing the brightness or as complex as multi-composition. In this paper, we consider the incorporation of meta-data to improve the construction of such graphs.}
\vspace{-10pt}   
\label{fig:teaser}
\end{figure}

The above case studies might lead one to believe that provenance analysis is a tool to decipher events far in the past. On the contrary, with the growth in popularity of online digital media, the need for provenance analysis has never been more timely. Current social sentiment can often only be fully understood within the context of online memes and other viral movements~\cite{fakenewsmexico}. Further, as the lines between real and fake images blur, the extent to which these types of online phenomena can be deployed towards the deception of the public has become deeply concerning~\cite{politifact}. With high quality cameras and image editing software at anybody's disposal, photographs have become easier to forge than paintings or sculptures. We have reached a point where digital forgeries can be produced with fine-grained detail, down to photographic style and sensor noise \cite{marra2014attacking,li2017anti}. These advancements in anti-forensics undermine the content's credibility, ownership, and authenticity. The current scale at which images and videos are shared requires an automated way of answering such questions.

%Unlike painting and sculptors, digital art can be produced very quickly in large numbers   
%In order to translate the techniques towards solving problems of image provenance, understanding the problem of provenance is crucial.
Image processing and computer vision techniques can be employed to detect correspondences between images or other digital art forms~\cite{Lowe:IJCV:2004,Bay:CVIU:2008,zagoruyko_2015}. This kind of correspondence can range from object matching in images~\cite{Lowe_1999} to comparing the style~\cite{gatys2015neural} and semantics~\cite{pan2004automatic} of the two. Provenance analysis can be thought of as ordering pair similarities between multiple image pair sets, and is therefore a natural extension to pairwise image comparison. These subsequent ordered parings can be modeled as a graph, where each edge denotes a correspondence between a pair, and the end vertices of the edge signify the two respective images. An example of such a graph can be seen in Figure~\ref{fig:teaser}.
This example shows that a provenance analysis algorithm could be analyzing multiple very close-looking realistic versions of the same visual object.
%Many types of different %types of
%photometric or geometric image operations are often applied to these images, further increasing the difficulty of matching the image's content. 
Complex scenarios like this can make content-based similarity metrics unreliable. 
% Depending on the types of transformation-invariance embedded in a model during learning the representations, multiple versions may map to the same space. In addition to that, the matching scenarios can become complex owing to the different types of transformations, belonging to a partially or fully unknown set, that could have been applied to the image beforehand. 

%% ------@DHM or @WJS - is the part below important?
%Among the large set of editing operations that must be accounted for, one of the most interesting and adversarial operations is splicing in foreign content to a photograph~\cite{ng2004model}. This results in an image that has multiple sources of original content but itself is not an original which complicates the process of ownership attribution and at times may also provide misleading information.  
%%-------------------------------------------

% Depending on the quality of the produced media object and the sensitivity of the topic addressed in it, it might be difficult as well as important to examine the composition of the multi-origin image.

\begin{figure}[t]
\begin{center}
   \includegraphics[width=1.0\linewidth]{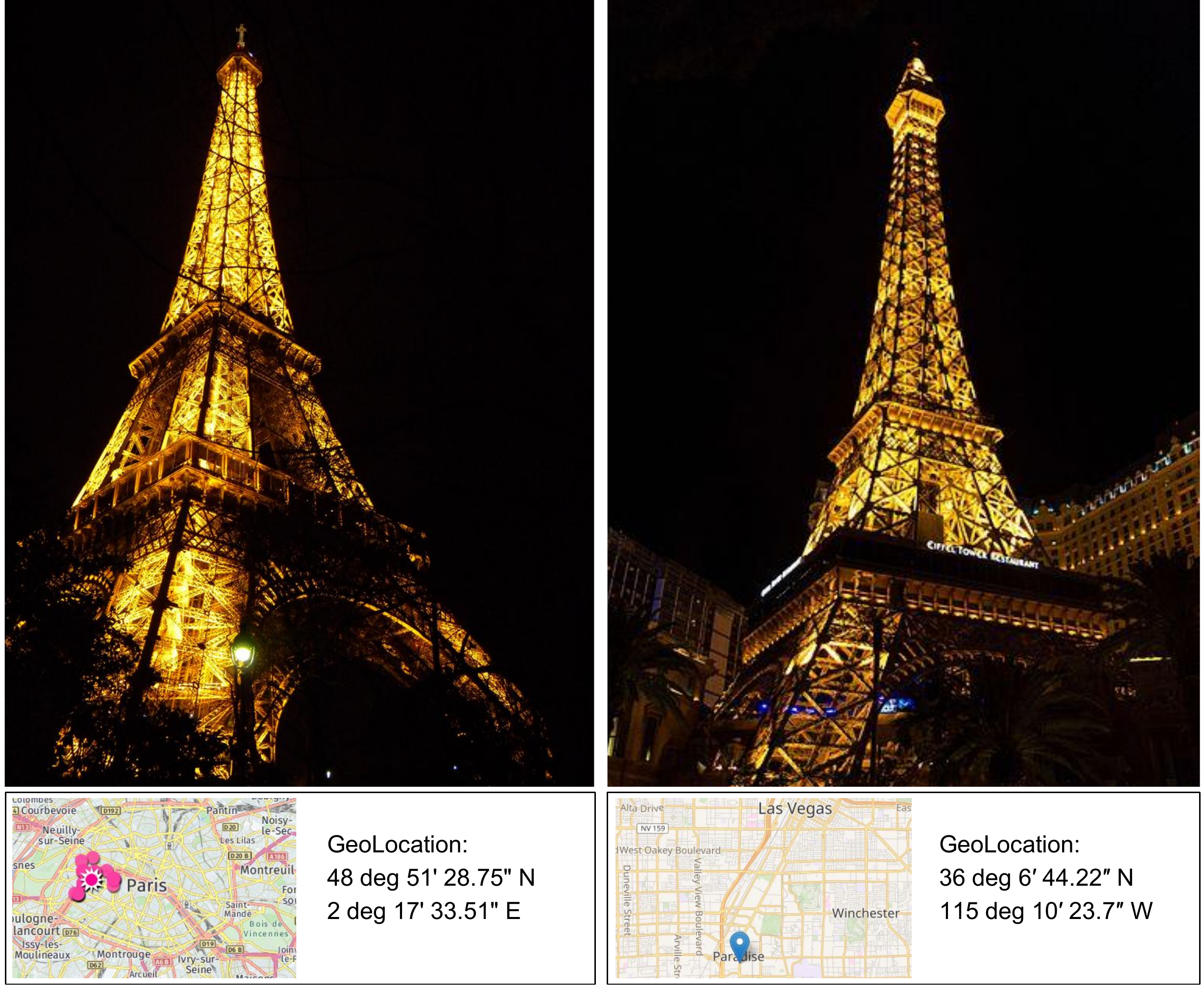}
\end{center}
   \caption{\textit{Left}: Photo of the Eiffel Tower taken at night in Paris. \textit{Right}: Photo of the replica of the monument in Las Vegas taken at night. Note that both photos depict the same visual object --- only the image file metadata in this case can help us understand that they are completely different scenes. Photos and their metadata were obtained from Flickr~\cite{flickreiffel} and Wikimedia Commons~\cite{wikieiffel}.}
\vspace{-10pt} 
\label{fig:diffloc}
\end{figure}

Due to the vast range of possible versions of a single original image, the metrics for quantifying the similarity between pairs of images can be noisy. Relying solely upon visual cues to order the different versions into a graph can result in poor provenance reconstructions~\cite{bharati2017uphy,moreira2018image}. Therefore, it becomes pertinent to utilize other sources of data to determine connections. For example, it is difficult to point out a semantic difference between the two images in Figure~\ref{fig:diffloc}, but the images can be differentiated by inspecting the metadata of the image files. Such a pair of images can be termed \textit{semantically similar}, as they are related to each other in a semantic way but do not originate from the same source ~\cite{oliveira2014multiple,bharati2017uphy}. Matching difficulty can also arise within sets of near-duplicate images, which are generated from a single origin having undergone a series of transformations (\textit{e.g.}, crop$\rightarrow$saturate$\rightarrow$desaturate). The pixel-level data within these image sets can exhibit ambiguous provenance directionality. Information beyond pixel-level data may be required to detect differences between such images.

To handle scenarios where image content fails to explain image evolution, file metadata can be used to help fill in the gaps. In this work, we explore the use of commonly present file metadata tags to improve image provenance analysis. We compare these results against image content-based methods and highlight the advantages and disadvantages of both. 

% A good solution should have the following characteristics:
% \begin{enumerate}
%     \item able to work across 
% \end{enumerate}

%-------------------------------------------------------------------------

\section{Related Work}
\label{sec:related}
Provenance analysis is a widely known and studied phenomenon in various data-based domains such as the semantic web and data warehousing~\cite{halaschek2006annotation, buneman2001and, anand2010provenance, simmhan2005survey}. However, provenance analysis for online multimedia has not been as extensively studied in the existing literature. 
% The methods discussed in the other-domain solutions cannot be applied directly to images due to their limitations and difference in the nature of data but the general principles are useful in understanding the nature of the problem.
The types of work most relevant and related to the problem of  image provenance analysis come from three established concepts in the digital forensics literature: near-duplicate detection~\cite{chum2008near, ke2004efficient}, image splicing detection~\cite{cozzolino2015splicebuster, bahrami2015blurred, iuliani2015image, chen2017image, huh2018fighting,Brogan2017Spotting} and image phylogeny~\cite{dias2012image, dias2013toward, de2016multiple}.
%Detecting near-duplicates is useful for efficient image search and retrieval by grouping near-duplicates together in image indexes and reducing storage use. 
%Other than increasing efficiency of large-scale image search engines, it can also be useful for detecting copyright infringement~\cite{ke2004efficient}. 
Most of the proposed methods work towards classifying whether an image is a near-duplicate of the query image in a retrieval context and do not determine the original image among the set of the near-duplicates. However, that particular problem has been studied by the image phylogeny community. 

Image phylogeny solutions aim at finding kinship relations between different versions of an image~\cite{dias2012image}. Similar to provenance analysis, image phylogeny limits its representation to a single-root tree with the original image as the root, even though there can be multiple original images contributing towards the creation of an image. The algorithm receives a query image and outputs the Image Phylogeny Tree (IPT). That method has also been extended to handle multiple (two) roots by taking spliced images into consideration~\cite{oliveira2014multiple}. 
%In a multiple parenting phylogeny pipeline, there are three types of images: host, alien and composite. The host is the base image, the alien image is the image donating the object to be spliced onto the base image and the composite image is the resulting spliced image. 
An example of this multiple parent scenario can be observed in Figure~\ref{fig:teaser} where four images (donors) contribute to the content of %one composite image.
 the central composite image.
%The problems tackled by these papers have covered a wide and complex scope of %problems
%situations, making the development of solutions difficult. For instance, differentiating between semantically similar images and near-duplicate images is an important task for all of these problems. Additionally, these algorithms must generalize across different forgery datasets, image transformations, file formats and image resolutions to be applicable in real world situations. Despite these requirements, the work discussed above solves very specific cases of image provenance analysis in the form of %research lab-
%synthetically generated forgeries, often breaking down when applied to unconstrained, real-world scenarios.
A constraint of these image phylogeny approaches that solve very specific cases of image provenance analysis is that they have dealt with constrained datasets using a limited set of transformations and image formats~\cite{jegou2008hamming,de2016multiple}. In addition to that, most of them only consider two images to form a composite, thereby limiting the solutions for large-scale general applicability. Thus new image provenance algorithms must generalize and be evaluated across different forgery datasets, image transformations, file formats and image resolutions to be applicable in real-world situations. 

As a step towards a more general framework for image provenance analysis inspired by image phylogeny works, recent work on undirected provenance graph construction~\cite{bharati2017uphy} 
adopted a more general taxonomy and dataset proposed by the American National Institute of Standards and Technology (NIST)~\cite{nist2017plan}. %proposed a more general taxonomy
It offered the \textit{U-phylogeny} pipeline as a preliminary approach towards solving provenance analysis, which is not restricted to either a closed set of image transformations, or the number of donor images to form multi-parent composites. 
%The U-phylogeny framework outputs results in the form of an undirected, acyclic graph (UDAG). 
% It is mostly designed and tested around small-scale provenance graphs belonging to the NIST 2016 Nimble Challenge dataset and only create UDAG provenance representations.
Results are presented for scenarios with and without the presence of distractors (images that are not related to the provenance history of the query image) showing the approach to be tolerant to irrelevant images. A limitation of the U-phylogeny approach is that it does not provide a directed provenance graph, which is required to understand the evolution of the media object. 

In order to overcome the direction limitation and propose a scalable approach, a more complete end-to-end pipeline for image provenance analysis was described in~\cite{moreira2018image}. That method for graph construction first builds dissimilarity matrices based on local image features, and then employs hierarchical clustering to group nodes and draw edges within the final provenance graph. As stated in Section~\ref{sec:intro}, relying solely on image content can lead to noisy edge inference. This is especially true for directed edges, which have been shown to be more difficult to derive than undirected edges~\cite{bharati2017uphy,moreira2018image}. An option for addressing this is the use of metadata related to the images. File metadata has been predominantly used for data and software provenance analysis~\cite{acar2010graph,anand2010provenance,halaschek2006annotation}, as such information reveals important clues about a file that cannot be directly derived from the data. Secondly, metadata related to online posts which include images can also be utilized for this purpose.

In the image domain, metadata often stores information regarding the device used to capture the image and the software used to process the image. Information provided by these types of tags has been utilized to improve the effectiveness of tasks such as image grouping~\cite{iqbal1999applying, logan2009automatic}, content-based image retrieval~\cite{akgul2011content, yee2003faceted}, photo classification~\cite{boutell2004photo}, image annotation~\cite{johnson2015love} and copyright protection~\cite{huang2010metadata}. Among these, algorithms establishing semantic correspondences between images, such as automatic grouping or classification, may utilize tags such as date, location, content originator, camera type and scene type~\cite{huiskes2008mir} whereas those that detect tampering may rely on detecting inconsistencies within the values of these and other tags containing source and copyright information~\cite{choi2013estimation, huang2010metadata}.
While metadata has been successfully used for forensics tasks in the past~\cite{birajdar2013digital, farid2009image,huh2018fighting, mahdian2010bibliography}, it has not been used for provenance analysis before.

\vspace{-8pt}
%---------------------------------------------------------
\section{Proposed Approach}
\label{sec:algo}

\begin{figure*}[t]
\begin{center}
   \includegraphics[width=1.0\linewidth]{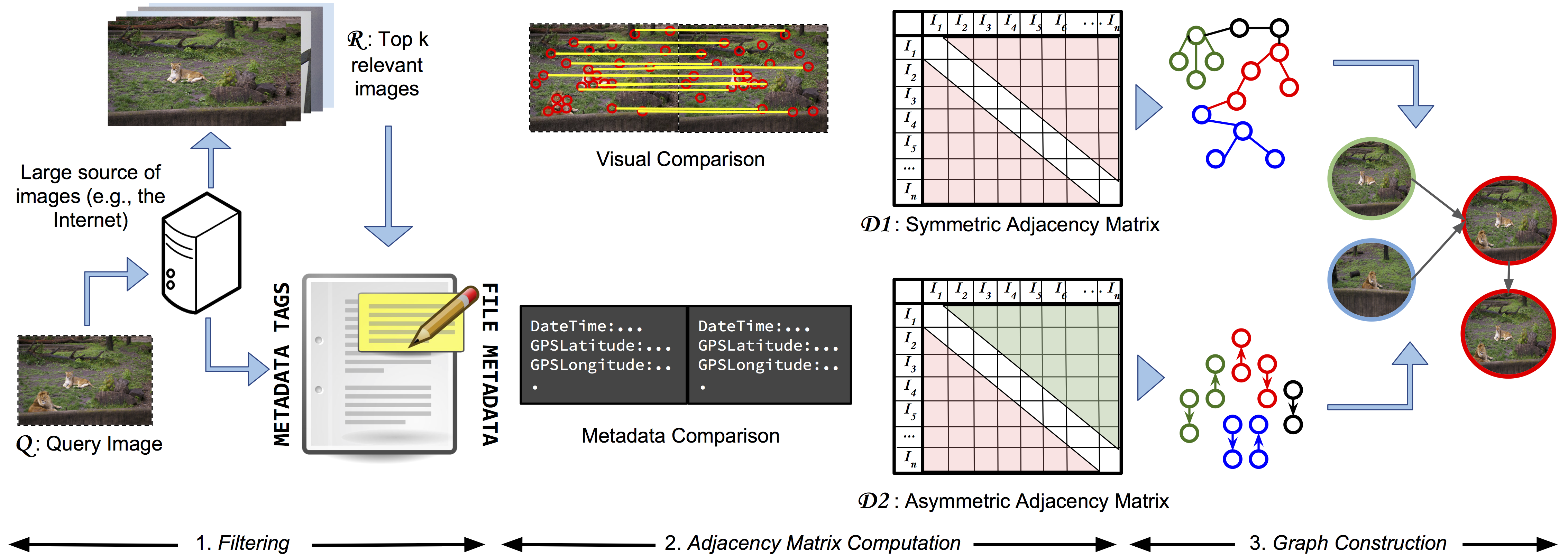}
\end{center}
   \caption{Stages of image provenance analysis. The proposed method starts with filtering images related to the provided query image $Q$. The `$k$' most relevant images are selected for pairwise image comparison. This step is not present in an oracle scenario where we assume to have been provided with the perfectly correct set of `$k$' related images. The images are compared in terms of visual content and metadata, yielding two types of adjacency matrices. The obtained matrices are then combined in the graph construction step to form an IPG.}
   \vspace{-2pt}
\label{fig:fullpipeline}
\end{figure*}

Image provenance analysis algorithms aim at constructing a provenance graph with related images, given a query image. The provenance graph~\cite{moreira2018image} is a Directed Acyclic Graph (DAG) where each node corresponds to an image in the set of related images and the edges stand for the relationship of sharing duplicate content. The direction of an edge denotes the direction of the flow of content between each pair of images and the overall provenance case. In this section, we explain in detail each of the three stages (as seen in Figure.~\ref{fig:fullpipeline}) of image provenance analysis used in the proposed approach.

%---------------------------------------------------------

%%------enter pipeline figure-----------------------------

%\subsection{{Retrieving the set of related images}}
\subsection{{Filtering}}
\label{subsec:filtering}
The first step required to perform provenance analysis for a given query image involves collecting the set of top-$k$ relevant images.
In this work, we follow the solution proposed in~\cite{moreira2018image}, which selects a subset of a large source of images (such as millions of images from the Internet), whose elements include samples sharing full content with the query with slight modifications (\textit{i.e.}, near-duplicate images), samples sharing partial content with the query in any form (single or multiple foreground objects, or background), and samples transitively related to the query (\textit{e.g.}, near duplicates of the images sharing content with the query).
In summary, this solution utilizes Optimized Product Quantization (OPQ) to store local Speeded-Up Robust Features (SURF)~\cite{Bay:CVIU:2008} in an Inverted File index (IVF), with a large number (\textit{e.g.},~$\sim$400k) of representative centroids.
Each image is described through at most 5k SURF features, which are fed to constitute the IVF index.
To search the index, multi-stage query expansion is utilized.
The first stage mainly retrieves the hosts, while the second stage retrieves donors; further stages retrieve the images transitively related to the query, by replacing the original query with samples retrieved in the previous stages.

%The first step required to perform provenance analysis for a given query image involves collecting the set of relevant images. The set of such images can be generated as a result of an image retrieval algorithm using the query. The algorithm, also known as ``Provenance Filtering", aims at selecting a subset of a large (typically, millions) set of images that share full content with the query image with slight modifications (\textit{i.e.}, near-duplicate images), ones that share partial content with the query image in any form (single or multiple foreground objects, or background), %, and their near-duplicates.
%and the images transitively related to the query (\textit{e.g.}, near duplicates of the images sharing content with the query). For the purpose of this work, we employ the technique proposed by Moreira et. al~\cite{moreira2018image}.  This technique utilizes Optimized Product Quantization (OPQ) to store local SURF features in an Inverted File index (IVF) with $\sim$400k centroids. Features extracted from the 5k points detected for each image are fed as input to the IVF index. To search the index, two stage query expansion is utilized. The first stage mainly works as retrieving the hosts while the second stage helps increase the donor image recall. 

%Upon receiving the set of images related to the query, the graph construction %can be
%is performed in two steps, respectively explained in Sections~\ref{sec:amc}~and~\ref{sec:gc}.

%\subsection{{Adjacency Matrix Creation}}
\subsection{{Adjacency Matrix Computation}}
\label{subsec:amc}
%\textbf{Creating a complete weighted adjacency matrix.}

Upon receiving the set of top-$k$ related images, denoted by $\mathcal{R}$, to the query image $Q$, we build $N \times N$ (here, $N=|\mathcal{R}|+1$) %asymmetric
adjacency matrices $\mathcal{D}$, in which each indexed value $\mathcal{D}[i,j]$ is the similarity (or dissimilarity) quotient between images $i$ and $j$. The full matrices are obtained by comparing %$\frac{n^2-n}{2}$
$(n^2-n) / 2$ pairs.

Different from previous work, though, besides using a matrix that relies solely on visual content, we propose the employment of an additional metadata-based asymmetric adjacency matrix that is used to determine the orientation of the pairwise image relations.
To the best of our knowledge, this is the first work proposing a way to leverage metadata to complement visual information for the problem of provenance analysis.
%Similarity values between pairs of images for the proposed approach have been obtained using both visual and metadata modalities. 

For visual comparison, the images can be described using interest point detectors and descriptors (such as SURF~\cite{Bay:CVIU:2008}) or learned from data using a Convolutional Neural Network. Image description for provenance analysis typically avoids using computationally expensive methods such as deep learning because of scalability concerns~\cite{zagoruyko_2015}. An empirical evaluation we conducted comparing SURF~\cite{Bay:CVIU:2008} and ShuffleNet~\cite{zhang2018shufflenet}, one of the most efficient deep learning frameworks, highlights this. Ignoring training time, ShuffleNet took 3.5 minutes to describe 10k images using two Nvidia Quadro GPUs, while SURF (a C++ implementation) took 39 seconds for the same images using one GPU. 
This motivates the usage of SURF-based detection and description of keypoints for the visual comparison between images. %point-based local description techniques for %provenance as their ultimate goal is to be deployed %at a large scale.
%The description is point of interest based and the points of interest are obtained using the SURF detector.
%\RED{After detecting and describing the $k$ most relevant interest points of each image,}
%Then, the obtained descriptors are matched using %the \textcolor{red}{FAST Matching or Brute-Force matching (check if there is a huge difference)}
Once the images are described, for each image pair, the $p$ most relevant interest points of each image are matched using brute-force pairwise comparison based on the L2 distance between the descriptors.
%Once described, the $p$ most relevant interest points of each image are then matched using brute-force pairwise comparison based on the L2 distance between the descriptors.
%The best matched correspondences are returned and top $k$ of them are chosen.
%Among the returned top matches, the ones that are geometrically inconsistent with the best matches are filtered out using the technique described in \cite{bharati2017uphy}.
The best matched correspondences are filtered to retain the geometrically consistent ones, as described in \cite{bharati2017uphy}. 
As a consequence, a symmetric adjacency matrix is obtained with the quantity of matched interest points between each pair of images. 

%The regions in the images including these points are then extracted as shared Regions of Interest (ROIs). Finally, the similarity between the pixel distributions in these regions is computed based on their mutual information which acts as a content-based relationship score between the images.
%A Region of Interest (ROI) for each image in the pair is obtained by fitting a rectangle around the convex hull of the set of matched interest points. The regions are masked out as the shared regions of interest.
%This phase of the matching or inverse dissimilarity computation deals with the estimation of structural correspondence between the images.
%In order to cover transformations that affect the texture (pixel intensities) without changing the geometry of the content in the image, we compute the mutual information between the two ROIs extracted for each pair of images.

%The proposed metadata-based solution builds upon the state-of-the-art provenance graph construction frameworks explained in the previous section. 

Commonly, the value of mutual information is used as the degree of pairwise association between images, or as asymmetric weights of the edges in a complete graph among the $N$ images, with no self loops \cite{moreira2018image}. In this work, in order to incorporate metadata information at this stage, we introduce a heuristic-based normalized voting to attribute weights to each pairwise image relationship. The voting method is chosen as a complement to the similarity comparison in the visual domain. The heuristics used to obtain the scores for each pair are %simple
straightforward metadata-related assumptions in the context of image provenance and rely upon the content of the tags. They include: %use the content of the tags  that are defined as follows:
\vspace{0.1cm}

\noindent \textbf{Date.} To check for the temporal order of content creation, we individually compare the date-related tags -- {\fontfamily{pcr}\selectfont DateTimeOriginal}, {\fontfamily{pcr}\selectfont ModifyDate} and {\fontfamily{pcr}\selectfont CreateDate}. Considering two images $i$ and $j$, for each one of the three dates, whenever available, the provenance relation $(i, j)$ gets one vote if the date of image $i$ is earlier or equal than the respective date of image $j$.
The relationship in the opposite direction $(j, i)$ is also analogously evaluated.
\vspace{0.1cm}

\noindent \textbf{Location.} Near-duplicates of an image (\textit{e.g.}, cropped versions) should have the same geographic location as the original one.
Hence, we cast one vote for the pairwise image relationship $(i, j)$, and one vote for the relationship $(j, i)$, if image $i$ shares with image $j$ exactly the same non-null values for the four location-related tags -- {\fontfamily{pcr}\selectfont GPSLatitude}, {\fontfamily{pcr}\selectfont GPSLatitudeRef}, {\fontfamily{pcr}\selectfont GPSLongitude}, {\fontfamily{pcr}\selectfont GPSLongitudeRef}.
Although this does not help to define the direction of the provenance between images $i$ and $j$, since both $(i, j)$ and $(j, i)$ relationships get one vote, it does help to give them more weight than the other image pairs that do not share location content.
In addition, in very complex image compositions where there is not a clear presence of a foreground donor, the location-related metadata tags might be null or missing, contrary to the donors of the composition. Thus, we alternatively cast one vote to the relationship $(i, j)$, if image $i$ has non-null location information and image $j$ is missing it.
\vspace{0.1cm}

\noindent \textbf{Camera.}
%Since modified versions of an image have the same source of imaging, a number of images can be discarded based on the same camera model check. 
We propose to use camera-based metadata information in a way that is analogous to the location case.
If image $i$ and image $j$ share the same non-null content for the camera's {\fontfamily{pcr}\selectfont Make}, {\fontfamily{pcr}\selectfont Model} and {\fontfamily{pcr}\selectfont Software} tags simultaneously, we cast one vote for both the $(i, j)$ and $(j, i)$ relationships, suggesting near-duplication that maintained image metadata.
Similarly, we cast one vote to $(i, j)$ if image $i$ has camera information and image $j$ does not.
\vspace{0.1cm}

\noindent \textbf{Editing.} 
%Some software packages such as Adobe Photoshop do leave traces of their usage in the metadata of the images through tags such as ImageResource and Software. The operations or modifications performed on the image may not be very drastic but the action gets registered.
We use the editing-related metadata tags to figure out if either image $i$ or image $j$ were ever manipulated.
Given that the provenance direction might occur from a non-manipulated to a manipulated image, we give one vote to the relationship $(i, j)$ if image $j$ has information for any of the {\fontfamily{pcr}\selectfont ProcessingSoftware}, {\fontfamily{pcr}\selectfont Artist}, {\fontfamily{pcr}\selectfont HostComputer}, {\fontfamily{pcr}\selectfont ImageResources}.
The relationship in the opposite direction $(j, i)$ is also evaluated in the same manner.
\vspace{0.1cm}

\noindent \textbf{Thumbnail.}
We extract the respective thumbnails of images $i$ and $j$.
If the thumbnails are exactly the same, both relationships $(i,j)$ and
$(j, i)$ get one vote, since it means one image might be generated from the other. %the images are related in terms of provenance.
Alternatively, if image $i$ has a thumbnail and image $j$ does not have one, then the relationship $(i, j)$ gets one vote, indicating that image $i$ is probably the original one.
\vspace{0.1cm}

These heuristics are used to generate a metadata-based image pairwise adjacency matrix $M$. 
For instance, taking images $i$ and $j$ and the possible provenance relationship from $i$ to $j$, whenever a heuristic is satisfied, the respective value $M[i,j]$ is increased in one unity, meaning the cast of one vote to the $(i, j)$ relationship.

\begin{figure}[t]
   \centering
   \includegraphics[width=1.0\linewidth]{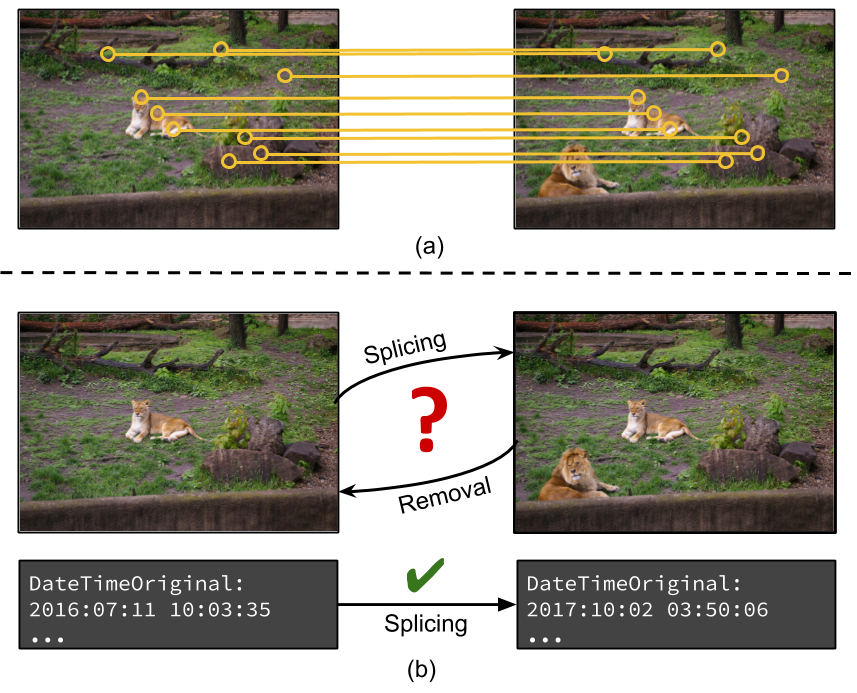}
   \caption{Usage of metadata information for determining direction in image pairwise provenance relationships.
   In (a), the output of interest-point-based analysis between two images is shown.
   The operation can be either a splicing or removal of the male lion.
   In (b), according to the date-based metadata, the operation is revealed to be a splice, since the image on the left is older.}
   \vspace{-10pt} 
\label{fig:algo}
\end{figure}

Aiming to keep the solution as widely applicable as possible, the tags are selected based on their availability and relevance to the provenance problem. An example of such relevance has been shown in Figure~\ref{fig:algo}. It depicts an image pair example that is directionally ambiguous. After performing interest-point-based pairwise analysis between the two images in Figure~\ref{fig:algo}(a), a valid argument for either a splicing (left-to-right edge) or removal (right-to-left edge) operation between the two could be made. Utilizing the ``DateTimeOriginal'' tag from both images disambiguates the relationship, revealing that the lion was indeed spliced into the image at a later time.
While a large array of metadata tags are often present in many images, only a small subset of these tags provide pertinent information useful for discerning inter-image relationships.
Furthermore, using tags provided by only specific camera firmwares or only applicable for certain formats (\eg, JPEG) reduces the generalizability of the proposed approach.
The tags mentioned here are EXIF tags (details provided in supplemental material) but the information provided by their values is what holds relevance to provenance. In case EXIF metadata is missing or tampered, information provided by online image posts, such as date of submission, can also be utilized in a similar way.

%%An additional aspect that we consider is the volatility of the information of a particular tag.
%%For example, tags such as TODO that TODO~\cite{exiv2} and is deemed to change in every TODO. 
%For example, generic tags such as {\fontfamily{pcr}\selectfont Exif.Image.ImageHistory} or tags with personally identifiable information such as {\fontfamily{pcr}\selectfont Exif.Image.Artist} or {\fontfamily{pcr}\selectfont Exif.Image.HostComputer} that are very useful for revealing tampering for image forensics but can be very prone to deletion or manipulation due to the privacy concerns or the malicious intent of the user.
%The availability and robustness qualities are related for some tags.

%Table~\ref{tab:tags} presents the metadata tags selected in this work for providing image pairwise provenance analysis.

%% ------------------------------------------------------------------

%\textcolor{red}{causation and correlation}

%\subsection{{DAG Construction}}
\subsection{{Graph Construction}}
\label{subsec:gc}    
%\textbf{Constructing the provenance graph.}
%Based on the comparative values of the metric,

Based on the values of the adjacency matrix, the final graph construction step chooses the most feasible set of directed edges (\ie, the set of edges that best represents the sequence of image operations).
%Based on the comparative values of the adjacency matrix, the final graph construction step chooses the most feasible (in terms of \RED{real-world} occurrences and order of image operations) $\theta(n)$ from the $(n^2-n)$ directed edges.
%that do not create a cycle. 
Each chosen directed edge denotes a parent-child relationship in the graph.  

Once the vision-based and metadata-based adjacency matrices are available, one can either individually use them to directly generate a provenance graph, through, for example, the application of Kruskal's Maximum Spanning Tree (MST) algorithm~\cite{Kruskal:AMS:1956}, or, as we are proposing, %augment the scores in the matrix generated by visual content using metadata-based scores before graph construction.
%In addition to the different styles of combining the adjacency matrices, the graph construction step can also vary.
%Upon obtaining the complete dissimilarity (or similarity) matrix, any minimum (or maximum) cost spanning tree algorithm can be used to build an undirected graph connecting relevant images and then, post-processed for directionality.
%Alternatively, 
%or
use a specialized algorithm for constructing a directed provenance graph, such as \emph{clustered provenance graph expansion}, proposed in~\cite{moreira2018image}. %., can be employed.
In the latter case, we suggest using the metadata-based asymmetric adjacency matrix to determine the directions of the edges.
In the experiments herein reported, we investigate both strategies.
%The method requires two inputs: (1) the number of retrieved images, and (2) the weighted adjacency matrix containing real-valued finite weight for each edge.
%The output graph is a binary adjacency matrix (BAM) for which BAM$[i,j]$ is set to 1 whenever there is an edge (\textit{i.e.}, $edge(i, j) \in$ MST).
In the end, the output graph can be represented as a binary adjacency matrix (BAM). BAM$[i,j]$ is set to 1 whenever there is an edge between images $i$ and $j$, indicating $i \rightarrow j$ flow of content.

Understandably, none of the proposed rules guarantee correct inference as metadata can be manipulated, wrong or missing. Using multiple tags reduces the impact of an incorrect inference and makes the process more robust. To mitigate circumstances where file metadata is unavailable, we demonstrate provenance in an online setting in our experiments using an alternative approach that can harvest metadata from website users' comments as opposed to the file itself. In both scenarios, the proposed approach is designed to tolerate events such as data tampering. As the metadata-compliance score is a cumulative score metric, each rule and the corresponding tags contribute to the value used to make the edge decision. 
% Using both metadata and harvested web data for this analysis establishes the robustness of the metric in cases where metadata is missing or tampered.

%------------------------------------------------------------------------
\vspace{-5pt}
\section{Experimental Setup}
\label{sec:exp}
\vspace{-5pt}
%Mention why 5k points of description and matches. Reason 1- Image is large so smaller numbers might not cut it. Reason 2- The number was empirically chosen after running filtering experiments for 1k, 2k, 5k and 10k.
%For the purpose of evaluating the proposed method, we create provenance graphs for two scenarios. The scenarios are analogous to the nature of the datasets, their method of curation and available useful metadata. In the sections below, we explain the characteristics and relevance of both of the datasets and other details of the setup. 
%We try the proposed method on two different datasets.
Here we detail the two evaluation scenarios and describe the characteristics of the corresponding datasets. 

\subsection{Provenance Analysis for Digital Forensics}
\label{subsec:nistprov}

NIST has recently released a dataset curated for the tasks of provenance image filtering and graph construction in a forensics context, which is devoid of most of the limitations of the existing datasets. 
Similar to the experimental setup described in~\cite{moreira2018image}, we rely on the \emph{development} partition of this dataset since it provides a full set of ground-truth graphs.
Named \emph{NC2017-Dev1-Beta4}, the dataset contains 65 queries, and the ground-truth is released in the form of journals depicting provenance graphs.
The provenance graph journals were created manually with the help of a proprietary image-editing journaling tool.
The graphs include links corresponding to simple image transformations such as cropping, scaling, sharpening, blurring, and rotation, to complex ones such as splicing from multiple sources and object removal.
The total number of related images per case ranges from $[2,81]$.
In addition to the images relevant to the provenance of each of the query images, the dataset also contains %nearly ten thousand
distractors (\textit{i.e.}, images not related to any query).

Following the protocol proposed by NIST~\cite{nist2017dataset}, we perform both \emph{end-to-end} and \emph{oracle-filter} provenance analysis over this dataset.
End-to-end analysis requires performing provenance filtering prior to graph construction \cite{pintoFiltering}.
In this case, for each query image, graphs are built upon a list of ranked images that might include distractors and miss genuinely related images due to imperfect image filtering.
To obtain these filtered image rank lists, we employ the best solution proposed in~\cite{moreira2018image} and retrieve the top-100 ranked images to the query, which may contain unrelated distractors.
Conversely, the oracle analysis does not require a filtering step, but instead starts with perfect ranks, \textit{i.e.}, ranks containing all the relevant images and no distractors.

Orthogonal to the \emph{end-to-end} versus \emph{oracle} comparison, we also compare results for both \emph{metadata only} and \emph{visual + metadata} solutions.
When using only metadata, we compute the vote-based metadata adjacency matrix, as explained in Section~\ref{subsec:amc}.
We use ExifTool~\cite{exiftool} to perform file metadata extraction. A table listing the tags used and their details has been provided in the supplemental material.
Once the adjacency matrix is computed, we apply Kruskal's maximum spanning tree algorithm~\cite{Kruskal:AMS:1956} to obtain the final provenance graph.

For fused metadata and visual solutions, we start with visual content-based adjacency matrices, which are generated according to the method explained in Section~\ref{sec:algo}.
We perform two different computations, one based on SURF~\cite{Bay:CVIU:2008} and the other based on Maximally Stable Extremal Regions (MSER)~\cite{Matas_2004}.
Both solutions were proposed and evaluated in~\cite{moreira2018image}, hence we follow their pipeline: (1) extraction of at most $5k$ interest points (either with SURF or MSER), (2) computation of adjacency matrices based on the number of geometrically consistent interest-point matches, (3) computation of adjacency matrices based on mutual information, and (4) application of the cluster-based method for generating provenance graphs.
%For combining visual content and metadata, we propose a slight change in the clustering algorithm: in the step of deciding the direction of edges: instead of immediately using the mutual-information-based adjacency matrices, we first consult the metadata-based ones.
%In this case, the direction of the edges is determined preferably by the metadata information.
For combining visual content and metadata, we proceed as suggested in Section~\ref{sec:algo}: % and perform a slight change in the cluster-based algorithm:
within the cluster-based algorithm, in the step of establishing the directions of edges, instead of using the mutual-information-based adjacency matrix~\cite{moreira2018image}, we consider the metadata-based one and keep the directions with more votes.
%For instance, given two images $i$ and $j$, and their respective metadata relationships $(i,j)$ and $(j,i)$ within the adjacency matrix, we keep, in the final provenance graph, the edge with highest value.}

\begin{table*}[t]
\renewcommand{\arraystretch}{1.2}
\caption{Results of provenance graph construction over the NIST NC2017-Dev1-Beta4 dataset.
We report the mean and the standard deviation for the metrics on the provided 65 queries.
Visual results are from Moreira et al.~\cite{moreira2018image}.
Best results are in bold.}
\centering
\footnotesize
\vspace{0.3cm}
\begin{tabular}{L{2.2cm}L{2.2cm}C{1.6cm}C{1.6cm}C{1.6cm}C{1.6cm}C{1.6cm}C{1.6cm}}
\hline
\multirow{2}{*}{\textbf{Data Modality}} & \multirow{2}{*}{\textbf{Solution}} & \multicolumn{3}{c}{\textbf{Oracle Filtering}} & \multicolumn{3}{c}{\textbf{End-to-End Analysis}}\\
\cmidrule(lr){3-5} \cmidrule(lr){6-8}
 & & \textbf{VO} & \textbf{EO} & \textbf{VEO} & \textbf{VO} & \textbf{EO} & \textbf{VEO} \\ % & \RED{Time (min)}\\
\hline
\multirow{2}{*}{Visual~\cite{moreira2018image}} & Cluster-SURF & 0.931$\pm$0.075 & 0.124$\pm$0.166 & 0.546$\pm$0.096 & 0.853$\pm$0.157 & 0.353$\pm$0.236 & 0.613$\pm$0.163\\ % & \RED{01.51}\\
 & Cluster-MSER & 0.892$\pm$0.154 & 0.123$\pm$0.161 & 0.525$\pm$0.129 & 0.835$\pm$0.180 & 0.312$\pm$0.252 & 0.585$\pm$0.177\\ % & \RED{01.35}\\
\cmidrule(lr){1-8}
\multirow{1}{*}{Metadata} & Kruskal & 0.999$\pm$0.003 & 0.117$\pm$0.099 & 0.577$\pm$0.053 & 0.249$\pm$0.115 & 0.009$\pm$0.016 & 0.130$\pm$0.057\\ % & \RED{\textbf{01.51}}\\
\cmidrule(lr){1-8}
\multirow{2}{*}{Visual + Metadata} & Cluster-SURF & \textbf{0.931$\pm$0.075} & \textbf{0.445$\pm$0.266} & \textbf{0.699$\pm$0.148} & \textbf{0.853$\pm$0.157} & \textbf{0.384$\pm$0.248} & \textbf{0.628$\pm$0.169} \\ % & \RED{\textbf{01.51}}\\
 & Cluster-MSER & 0.891$\pm$0.154 & 0.389$\pm$0.254 & 0.651$\pm$0.176 & 0.838$\pm$0.182 & 0.345$\pm$0.232 & 0.603$\pm$0.174\\ % & \RED{\textbf{01.51}}\\
\hline
\end{tabular}
\vspace{-10pt}
\label{tab:nist}
\end{table*}

The provenance graphs generated using the proposed approach for both oracle and end-to-end scenarios are evaluated using the metrics proposed by NIST for the provenance task~\cite{nist2017plan}. The metrics focus on comparing the nodes and edges from both ground-truth and candidate graphs. The corresponding measures of  \textbf{Vertex Overlap (VO)} and \textbf{Edge Overlap (EO)} are the harmonic mean of precision and recall (F1 score) for the nodes and edges retrieved by our method. In addition to these, a unified metric representing one score for the graph overlap namely the \textbf{Vertex Edge Overlap (VEO)} is also reported. The VEO is the combined F1 score for nodes and edges. All the metrics are computed through the NIST \emph{MediScore} tool~\cite{mediscore}. 
%The tool allows for the evaluation of multiple tasks but we choose the provenance graph building option for our purpose.
The values of these metrics lie in the range $[0,1]$ where higher values are better. 
%The three scores are computed for each query or provenance case and the average value of these over all the cases present in the dataset is reported.

\subsection{Provenance Analysis for Cultural Analytics}
\label{subsec:reddit}

To include experiments with more realistic examples, we also evaluate the approaches from Section \ref{sec:algo} on the Reddit dataset introduced in~\cite{moreira2018image} and maintained at~\cite{redditdataset}.
This dataset contains provenance cases created from images extracted from the \emph{photoshopbattles} community on the Reddit website~\cite{reddit2017photoshopbattles}.
This community provides a platform for users to experiment with image manipulation in a friendly context.
Each thread begins with a single image submitted by one user, which serves as the base image for the manipulations of others, whose contributions appear as comments on the original post.
For the purpose of provenance, Moreira et  al.~\cite{moreira2018image} utilize this comment structure to obtain 184 provenance graphs with an average graph order of 56. For the sake of fair comparison, we evaluate the variants of the proposed approach on the exact same set. The full set of images from Reddit do not contain distractors.
This restricts our experiments for provenance analysis in this setting to \emph{oracle-filter} analysis only, in contrast to the NC2017-Dev1-Beta4 dataset.

%Since the images in this dataset are collected from the web, the availability of metadata is restricted by the policies and approaches of the Reddit website and image hosting websites such as imgur.com~\cite{imgur2018privacy}. The metadata tags available with online hosted memes and other images are related to the web upload and there are not many tags directly involving image properties and history. On the other hand, there can be post related external information that can support provenance.

%Even though there are other image hosting websites such Flickr, Picasa that preserve more of the metadata tags, they do not provide a structured information for provenance ground truth and hence, cannot be utilized for evaluation for provenance graphs. Nevertheless, the datasets and the available tags have found wide usage for evaluation of image retrieval solutions~\cite{huiskes2008mir} and have supported other computer vision applications~\cite{mcauley2012image}. We evaluate the provenance approaches in the Reddit setting, to test the feasibility of current approaches in a specific provenance setting on an online platform. For conducting provenance on the this dataset, we utilize the dates collected from Reddit posts and comments. We run experiments to match the oracle scenario of NIST and present the results for all the three experiments based on the modality used to establish the edges -- image only, metadata only and both. The same metrics and scorer as the NIST dataset is used to evaluate these graphs as well. 

Since the images in the Reddit dataset are collected from the web, the availability of metadata is restricted by the policies of the Reddit website and image hosting services, such as \url{imgur.com}~\cite{imgur2018privacy}.
For that reason, the metadata extraction through ExifTool~\cite{exiftool} does not deliver useful tags for provenance analysis.
As an alternative, we use the Reddit users' comments and posts to estimate the date and time of image uploads, thus treating them as \emph{DateTimeOriginal} values, making it possible to invoke the date-based heuristics.

Here, one important comment can be made about the restricted availability of metadata and the apparent limited possibility of application of the present solution.
Although metadata might not be available to the general public, image hosting websites might still be storing them, hence being able to apply the method in their headquarters or under legal demand.
Other image hosting websites such as Flickr and Picasa can be used as image sources that preserve metadata tags, but they do not provide structured information for provenance ground-truth extraction.
%, nor they are primary outlets for meme-style imagery. 
This promotes Reddit as a choice for obtaining graphs and evaluating provenance in a cultural setting.
To evaluate our experimental results on the Reddit dataset, we employ the same metrics and scorer used in the case of the NC2017-Dev1-Beta4 dataset.

%%-------------------------------------------------------------------------
\begin{table*}[t]
\renewcommand{\arraystretch}{1.1}
\caption{Ablation results for oracle and end-to-end provenance.
We repeat the experiments seven times for the best solution presented in Table~\ref{tab:nist} (Visual + metadata, Cluster-SURF) in both scenarios, keeping only a subset of heuristics activated at a time.
Best results in bold.}
\centering
\footnotesize
\vspace{0.3cm}
\begin{tabular}{R{2.5cm}C{2cm}C{2cm}C{2cm}C{2cm}C{2cm}C{2cm}}
\hline
\multirow{2}{*}{\textbf{Heuristic}} & \multicolumn{3}{c}{\textbf{Oracle Filtering}} & \multicolumn{3}{c}{\textbf{\RED{End-to-End Analysis}}}\\
\cmidrule(lr){2-4} \cmidrule(lr){5-7}
 & \textbf{VO} & \textbf{EO} & \textbf{VEO} & \RED{\textbf{VO}} & \RED{\textbf{EO}} & \RED{\textbf{VEO}} \\ % & \RED{Time (min)}\\
\hline
       Date only & \textbf{0.931$\pm$0.075} & \textbf{0.446$\pm$0.265} & \textbf{0.700$\pm$0.147} & \RED{0.853$\pm$0.157} & \RED{0.389$\pm$0.244} & \RED{0.630$\pm$0.169} \\
   Location only & 0.931$\pm$0.075 & 0.394$\pm$0.282 & 0.674$\pm$0.154 & \RED{0.853$\pm$0.157} & \RED{0.348$\pm$0.241} & \RED{0.611$\pm$0.164} \\
     Camera only & 0.931$\pm$0.075 & 0.388$\pm$0.269 & 0.672$\pm$0.147 & \RED{0.853$\pm$0.157} & \RED{0.350$\pm$0.234} & \RED{0.612$\pm$0.164} \\
    Editing only & 0.931$\pm$0.075 & 0.396$\pm$0.281 & 0.675$\pm$0.153 & \RED{0.853$\pm$0.157} & \RED{0.353$\pm$0.237} & \RED{0.613$\pm$0.163} \\
  Thumbnail only & 0.931$\pm$0.075 & 0.411$\pm$0.285 & 0.683$\pm$0.155 & \RED{0.853$\pm$0.157} & \RED{0.363$\pm$0.238} & \RED{0.618$\pm$0.167} \\
   \RED{ All but Date} & \RED{0.931$\pm$0.075} & \RED{0.394$\pm$0.280} & \RED{0.675$\pm$0.152} & \RED{0.853$\pm$0.157} & \RED{0.345$\pm$0.247} & \RED{0.610$\pm$0.168} \\
\RED{Date + Thumbnail} & \RED{0.931$\pm$0.075} & \RED{0.444$\pm$0.268} & \RED{0.699$\pm$0.148} & \RED{\textbf{0.853$\pm$0.157}} & \RED{\textbf{0.391$\pm$0.245}} & \RED{\textbf{0.632$\pm$0.169}} \\
\hline
\end{tabular}
\vspace{-8pt}
\label{tab:ablat}
\end{table*}

\section{Experimental Results}
\label{subsec:results}

The experiments performed on both datasets show that utilizing knowledge from metadata helps in the process of edge inference for provenance. As it can be observed from the values reported in Table~\ref{tab:nist}, the proposed method significantly improves total edge overlap, and thereby total graph overlap, since it uses image-content-based information to initially establish connections between images, then relies on metadata to refine edge direction. The tags and checks used in this work yield an edge overlap of 44.5\% and graph overlap (VEO) of $\sim$70\% for provenance in the oracle scenario, improving notably over current state-of-the-art~\cite{moreira2018image} by %$\sim$15\%. 
$\sim$15 percentage points (pp). More notably, metadata fusion provides a %$\sim$30\% 
$\sim$30pp increase in EO in the oracle cases, when compared to~\cite{moreira2018image}.

In the end-to-end scenario, metadata usage also shows improvements in edge overlap by %3-3.5\% 
$\sim$3pp, aiding the overall graph overlap to reach $>$60\%. Provenance analysis solutions thus far have struggled at obtaining good edge reconstruction, as can be seen from the disparity between the vertex and edge overlap. Furthermore, the addition of distractors reduces performance by %$\sim$5\%
$\sim$5pp, implying that semantically similar images within the distractor sets can lead to high inter-image similarity between pairs that should not be related. This can negatively impact greedy graph construction approaches.
Some success and failure provenance cases are presented %in Figure~\ref{fig:output} and
in the supplemental material, including the graph visualizations.

To understand the contribution of each type of metadata information, we conduct an ablation study on the oracle and end-to-end scenarios using the \emph{Visual+Metadata, Cluster-SURF} method from Section~\ref{subsec:nistprov}.
We perform the experiment seven times, for each scenario, using only a subset of heuristics for each run.
Results are presented in Table~\ref{tab:ablat}.
In the oracle scenario, while all five tags individually benefit graph EO, the date-based one performs best, followed by thumbnail usage.
For that reason, we also present, in the last two rows of Table~\ref{tab:ablat}, the results of having all heuristics combined except for date (to assess the impact of avoiding the best one), as well as combination of date and thumbnail (the two best ones).
Indeed, the date-based heuristic alone slightly surpasses the combination of heuristics, in this particular dataset and scenario.
In the end-to-end scenario, in turn, observations are somewhat different.
Metadata tags alone do not improve the results of the visual solution, except for date and the date-thumbnail fusion, with the latter showing the best results.
Again, this might be particular to the dataset, where the added distractors probably present more unreliable metadata (due to tampering or removal).
That reveals the importance of combining tags, since it leads to a more robust solution to metadata tampering.

\begin{table}[t]
\renewcommand{\arraystretch}{1.1}
\caption{Results of provenance graph construction over the Reddit dataset.
We report the average values of the metrics over the 184 cases, as well as the standard deviations.
This dataset only allows us to report  oracle-filtering results.
Visual results are from Moreira et al.~\cite{moreira2018image}.
Best results are in bold.
}
%Combined implies a fused solution with both SURF and MSER. 
% In bold, the solutions with the best VEO.}
\centering
\footnotesize
\vspace{0.3cm}
\begin{tabular}{R{2cm}C{1.5cm}C{1.5cm}C{1.5cm}}
% % \begin{tabular}{C{1.0cm}R{1.1cm}C{1cm}C{1cm}C{1cm}}
\hline
\multicolumn{1}{c}{\textbf{Solution}} & \textbf{VO} & \textbf{EO} & \textbf{VEO}\\ % & \RED{Time (min)}\\
\hline
\multicolumn{4}{l}{Visual~\cite{moreira2018image}:} \\
\ \ \ \ Cluster-SURF & 0.757$\pm$0.341 & 0.037$\pm$0.034 & 0.401$\pm$0.181\\ % & \RED{\textbf{11.71}}\\
\ \ \ \ Cluster-MSER & 0.509$\pm$0.388 & 0.027$\pm$0.034 & 0.271$\pm$0.207 \\
\cmidrule(lr){1-4}
\multicolumn{4}{l}{Metadata:} \\
\ \ \ \ Kruskal & \textbf{0.969$\pm$0.073} & 0.034$\pm$0.086 & \textbf{0.506$\pm$0.056} \\ % &\RED{\textbf{01.51}}\\
\cmidrule(lr){1-4}
\multicolumn{4}{l}{Visual + Metadata:} \\
\ \ \ \ Cluster-SURF & 0.757$\pm$0.341 & \textbf{0.085$\pm$0.065} & 0.424$\pm$0.193 \\ % & \RED{\textbf{01.51}}\\
\ \ \ \ Cluster-MSER & 0.509$\pm$0.388 & 0.061$\pm$0.063 & 0.288$\pm$0.220 \\
\hline
\end{tabular}
\label{tab:reddit}
\end{table}

% %-------------------------------------------------------------------------
Provenance analysis becomes significantly more difficult when dealing with real-world scenarios, such as those presented in the Reddit dataset. Although metadata doubles the number of correctly retrieved edges, as seen in Table~\ref{tab:reddit}, the edge overlap is still much lower than for the NC2017-Dev1-Beta4 dataset. In the Reddit cases, images can be connected by visual puns, inside jokes, and purely associative content without any direct visual correspondence between them. This is very common in meme-style imagery. Understanding the quirks and sentiments of human language can further help provenance analysis in these contexts, but it has not yet been explored. To perform complex relationship retrieval using image provenance analysis, input from other modalities, such as text comments, may be required.

Since all experiments calculate initial correspondences using only visual image content, the purely visual method and visual + metadata based method perform identically with respect to VO. This metric is generally high with a low standard deviation whereas the EO has very high standard deviation. Due to the vast range of possible transformations, the provenance analysis approaches are not able to detect and map certain image relationships as well as others. The results of the experiments for both scenarios show that SURF detections for image matching are better than MSER detections, which is consistent with the results in~\cite{moreira2018image}.

% \begin{figure}[t]
%   \centering
%   \includegraphics[width=1.0\linewidth]{figures/sampleoutput5.pdf}
%   \caption{Visualizations of partial provenance graphs constructed for a case in the %oracle
%   end-to-end scenario (NC2017-Dev1-Beta4) using the Cluster-SURF method.
%   The left graph generation uses only image content while the right one utilizes metadata. This is one of the cases where adding metadata %drastically
%   improves edge inference.
%   In the left-hand graph %all of the edges
%   two of the edges are wrong (dashed), while the use of metadata has corrected %all but one of them in the right-hand graph.
%   them, in the right-hand graph. %The corresponding correct edge is denoted by the black edge.
%   The complete graph for %the end-to-end scenario
%   this case is presented in the supplemental material.}
% \label{fig:output}
% \end{figure}

% %%------------------------------------------------------------------------
\section{Discussion}
\label{sec:discussion}

Image metadata is a valuable asset for improving results for vision-based problems such as image retrieval~\cite{sasikala2015efficient}, semantic segmentation~\cite{ardeshir2015geo}, and manipulation detection~\cite{huh2018fighting}. Our work demonstrates that the task of image provenance analysis also benefits from metadata. External context can corroborate evidence from purely visual techniques, creating an overall better solution to provenance graph reconstruction. 

In addition to utilizing information that cannot be derived from the images themselves, metadata-based approaches are computationally very cheap. Furthermore, unlike complex, data-driven, vision-based techniques that require large amounts of training resources, methods like ours require no training at all. Such methods can be deployed easily on a large scale, incurring very little performance overhead. Approaches that require large amounts of training data can suffer due to the relatively small sizes of currently available provenance datasets. And most datasets published in this field so far are indeed small.

Even though external information can improve image-based approaches, provenance analysis is still far from being solved. This work only presents a preliminary exploration of utilizing metadata in provenance analysis. While our results show improvement, metadata-based approaches have higher chances of being rendered unreliable due to their absence or manipulation. Further advancements in solving the problem must focus on the examination of content-derived metadata as well. Future work could include estimating missing metadata information from the content and available tags~\cite{fan2013estimating,tsai2005extent}. For now, our findings suggest that image-content-based methods should be the fallback option, as metadata alone is more useful for determining edge directions instead of edge selection. We surmise that going forward, the best provenance approaches should rely primarily on image content, but utilize metadata analysis as a secondary refinement system in scenarios where it is present and provides ample evidence. 

%Solutions can only utilize information in cases that 

%Discuss about the reliability of metadata.
%For cases, where we don't have metadata available, there are also methods that can be used to predict/estimate metadata~\cite{fan2013estimating,tsai2005extent}

%Also, advantages of this approach - learning free, computationally less expensive, 

%------------------------------------------------------------------------
% There is no need for a conclusion section - WJS
%\section{Conclusion}
%\label{sec:conclusion}

%---------------------------------------------------------

{\small
\bibliographystyle{ieee}
\bibliography{egbib}

\begin{thebibliography}{10}\itemsep=-1pt

\bibitem{acar2010graph}
U.~Acar, P.~Buneman, J.~Cheney, J.~Van Den~Bussche, N.~Kwasnikowska, and
  S.~Vansummeren.
\newblock A graph model of data and workflow provenance.
\newblock In {\em USENIX Workshop on the Theory and Practice of Provenance},
  2010.

\bibitem{akgul2011content}
C.~B. Akg{\"u}l, D.~L. Rubin, S.~Napel, C.~F. Beaulieu, H.~Greenspan, and
  B.~Acar.
\newblock Content-based image retrieval in radiology: current status and future
  directions.
\newblock {\em Springer Journal of Digital Imaging}, 24(2):208--222, 2011.

\bibitem{alvarez2004using}
P.~Alvarez.
\newblock Using extended file information (exif) file headers in digital
  evidence analysis.
\newblock {\em International Journal of Digital Evidence}, 2, 2004.

\bibitem{anand2010provenance}
M.~K. Anand, S.~Bowers, and B.~Lud{\"a}scher.
\newblock Provenance browser: Displaying and querying scientific workflow
  provenance graphs.
\newblock In {\em IEEE International Conference on Data Engineering}, pages
  1201--1204, 2010.

\bibitem{ardeshir2015geo}
S.~Ardeshir, K.~Malcolm Collins-Sibley, and M.~Shah.
\newblock Geo-semantic segmentation.
\newblock In {\em IEEE Conference on Computer Vision and Pattern Recognition},
  pages 2792--2799, 2015.

\bibitem{bahrami2015blurred}
K.~Bahrami, A.~C. Kot, L.~Li, and H.~Li.
\newblock Blurred image splicing localization by exposing blur type
  inconsistency.
\newblock {\em IEEE Transactions on Information Forensics and Security},
  10(5):999--1009, 2015.

\bibitem{Bay:CVIU:2008}
H.~Bay, A.~Ess, T.~Tuytelaars, and L.~Van~Gool.
\newblock Speeded-up robust features ({SURF}).
\newblock {\em Computer Vision and Image Understanding}, 110(3):346--359, June
  2008.

\bibitem{bbcprogram}
{BBC One}.
\newblock {Fake or Fortune?}
\newblock \url{https://www.bbc.co.uk/programmes/b01mxxz6}.
\newblock Accessed on 06-27-2018.

\bibitem{fakeorfortune_2012}
{BBC One}.
\newblock {Fake or Fortune? Degas and the Little Dancer}.
\newblock \url{https://www.bbc.co.uk/programmes/p00xym5k}, 2012.

\bibitem{fakeorfortune_2015}
{BBC One}.
\newblock {Fake or Fortune? Lowry}.
\newblock \url{https://www.bbc.co.uk/programmes/p02whms0}, 2015.

\bibitem{bharati2017uphy}
A.~Bharati, D.~Moreira, A.~Pinto, J.~Brogan, K.~Bowyer, P.~Flynn, W.~Scheirer,
  and A.~Rocha.
\newblock U-phylogeny: Undirected provenance graph construction in the wild.
\newblock In {\em IEEE International Conference on Image Processing}, 2017.

\bibitem{birajdar2013digital}
G.~K. Birajdar and V.~H. Mankar.
\newblock Digital image forgery detection using passive techniques: A survey.
\newblock {\em Elsevier Digital Investigation}, 10(3):226--245, 2013.

\bibitem{flickreiffel}
J.~Bon.
\newblock {Paris Eiffel Tower}.
\newblock \url{https://www.flickr.com/photos/girolame/3220601379/}, 2009.
\newblock Accessed on 06-25-2018.

\bibitem{boutell2004photo}
M.~Boutell and J.~Luo.
\newblock Photo classification by integrating image content and camera
  metadata.
\newblock In {\em IEEE International Conference on Pattern Recognition},
  volume~4, pages 901--904. IEEE, 2004.

\bibitem{Brogan2017Spotting}
J.~Brogan, P.~Bestagini, A.~Bharati, A.~Pinto, D.~Moreira, K.~Bowyer, P.~Flynn,
  A.~Rocha, and W.~Scheirer.
\newblock Spotting the difference: Context retrieval and analysis for improved
  forgery detection and localization.
\newblock In {\em IEEE International Conference on Image Processing}, pages
  4078--4082, 2017.

\bibitem{buneman2001and}
P.~Buneman, S.~Khanna, and T.~Wang-Chiew.
\newblock Why and where: A characterization of data provenance.
\newblock In {\em Springer International conference on database theory}, pages
  316--330, 2001.

\bibitem{chen2017image}
C.~Chen, S.~McCloskey, and J.~Yu.
\newblock Image splicing detection via camera response function analysis.
\newblock In {\em IEEE Conference on Computer Vision and Pattern Recognition},
  pages 5087--5096, 2017.

\bibitem{choi2013estimation}
C.-H. Choi, H.-Y. Lee, and H.-K. Lee.
\newblock Estimation of color modification in digital images by cfa pattern
  change.
\newblock {\em Elsevier Forensic Science International}, 226(1-3):94--105,
  2013.

\bibitem{chum2008near}
O.~Chum, J.~Philbin, A.~Zisserman, et~al.
\newblock Near duplicate image detection: min-hash and tf-idf weighting.
\newblock In {\em British Machine Vision Conference}, volume 810, pages
  812--815, 2008.

\bibitem{redditdataset}
{Computer Vision Research Lab}.
\newblock {ND Reddit Provenance Dataset}.
\newblock \url{https://github.com/CVRL/Reddit_Provenance_Datasets}, 2018.
\newblock {Accessed on 07-02-2018}.

\bibitem{cozzolino2015splicebuster}
D.~Cozzolino, G.~Poggi, and L.~Verdoliva.
\newblock Splicebuster: A new blind image splicing detector.
\newblock In {\em IEEE International Workshop on Information Forensics and
  Security}, pages 1--6, 2015.

\bibitem{de2016multiple}
A.~A. de~Oliveira, P.~Ferrara, A.~De~Rosa, A.~Piva, M.~Barni, S.~Goldenstein,
  Z.~Dias, and A.~Rocha.
\newblock Multiple parenting phylogeny relationships in digital images.
\newblock {\em IEEE Transactions on Information Forensics and Security},
  11(2):328--343, 2016.

\bibitem{dias2013toward}
Z.~Dias, S.~Goldenstein, and A.~Rocha.
\newblock Toward image phylogeny forests: Automatically recovering semantically
  similar image relationships.
\newblock {\em Elsevier Forensic science international}, 231(1):178--189, 2013.

\bibitem{dias2012image}
Z.~Dias, A.~Rocha, and S.~Goldenstein.
\newblock Image phylogeny by minimal spanning trees.
\newblock {\em IEEE Transactions on Information Forensics and Security},
  7(2):774--788, 2012.

\bibitem{douglas2010origins}
J.~Douglas.
\newblock Origins: evolving ideas about the principle of provenance.
\newblock {\em Currents of Archival Thinking}, 24, 2010.

\bibitem{exiv2}
Exiv2.
\newblock Metadata reference tables.
\newblock \url{http://www.exiv2.org/metadata.html}, 2017.
\newblock Accessed on 06-28-2018.

\bibitem{fan2013estimating}
J.~Fan, H.~Cao, and A.~C. Kot.
\newblock Estimating exif parameters based on noise features for image
  manipulation detection.
\newblock {\em IEEE Transactions on Information Forensics and Security},
  8(4):608--618, 2013.

\bibitem{farid2009image}
H.~Farid.
\newblock Image forgery detection.
\newblock {\em IEEE Signal Processing Magazine}, 26(2):16--25, 2009.

\bibitem{gatys2015neural}
L.~A. Gatys, A.~S. Ecker, and M.~Bethge.
\newblock A neural algorithm of artistic style.
\newblock {\em arXiv preprint arXiv:1508.06576}, 2015.

\bibitem{halaschek2006annotation}
C.~Halaschek-Wiener, J.~Golbeck, A.~Schain, M.~Grove, B.~Parsia, and
  J.~Hendler.
\newblock Annotation and provenance tracking in semantic web photo libraries.
\newblock In {\em Springer International Provenance and Annotation Workshop},
  pages 82--89, 2006.

\bibitem{exiftool}
P.~Harvey.
\newblock Exiftool.
\newblock \url{https://www.sno.phy.queensu.ca/~phil/exiftool/}, 2018.
\newblock Accessed on 07-01-2018.

\bibitem{politifact}
A.~D. Holan.
\newblock 2016 lie of the year: Fake news.
\newblock
  \url{http://www.politifact.com/truth-o-meter/article/2016/dec/13/2016-lie-year-fake-news/},
  2016.
\newblock Accessed on 06-28-2018.

\bibitem{huang2010metadata}
H.-C. Huang and W.-C. Fang.
\newblock Metadata-based image watermarking for copyright protection.
\newblock {\em Elsevier Simulation Modelling Practice and Theory},
  18(4):436--445, 2010.

\bibitem{huh2018fighting}
M.~Huh, A.~Liu, A.~Owens, and A.~A. Efros.
\newblock Fighting fake news: Image splice detection via learned
  self-consistency.
\newblock {\em arXiv preprint arXiv:1805.04096}, 2018.

\bibitem{huiskes2008mir}
M.~J. Huiskes and M.~S. Lew.
\newblock The mir flickr retrieval evaluation.
\newblock In {\em ACM International Conference on Multimedia Information
  Retrieval}, pages 39--43, 2008.

\bibitem{imgur2018privacy}
{imgur.com}.
\newblock {Imgur Privacy Standards}.
\newblock
  \url{https://help.imgur.com/hc/en-us/articles/201746817-Post-privacy}, 2018.
\newblock Accessed on 06-25-2018.

\bibitem{iqbal1999applying}
Q.~Iqbal and J.~Aggarwall.
\newblock Applying perceptual grouping to content-based image retrieval:
  Building images.
\newblock In {\em IEEE Conference on Computer Vision and Pattern Recognition},
  volume~1, pages 42--48, 1999.

\bibitem{iuliani2015image}
M.~Iuliani, G.~Fabbri, and A.~Piva.
\newblock Image splicing detection based on general perspective constraints.
\newblock In {\em IEEE International Workshop on Information Forensics and
  Security (WIFS)}, pages 1--6, 2015.

\bibitem{jegou2008hamming}
H.~Jegou, M.~Douze, and C.~Schmid.
\newblock Hamming embedding and weak geometric consistency for large scale
  image search.
\newblock In {\em Springer European Conference on Computer Vision}, pages
  304--317, 2008.

\bibitem{johnson2015love}
J.~Johnson, L.~Ballan, and L.~Fei-Fei.
\newblock Love thy neighbors: Image annotation by exploiting image metadata.
\newblock In {\em IEEE International Conference on Computer Vision}, pages
  4624--4632, 2015.

\bibitem{ke2004efficient}
Y.~Ke, R.~Sukthankar, L.~Huston, Y.~Ke, and R.~Sukthankar.
\newblock Efficient near-duplicate detection and sub-image retrieval.
\newblock {\em ACM Multimedia}, 4(1):1--5, 2004.

\bibitem{kee2011digital}
E.~Kee, M.~K. Johnson, and H.~Farid.
\newblock Digital image authentication from jpeg headers.
\newblock {\em IEEE Transactions on Information Forensics and Security},
  6:1066--1075, 2011.

\bibitem{Kruskal:AMS:1956}
J.~Kruskal.
\newblock On the shortest spanning subtree of a graph and the traveling
  salesman problem.
\newblock {\em Proceedings of the American Mathematical Society}, 7(1):48--50,
  1956.

\bibitem{li2017anti}
H.~Li, W.~Luo, Q.~Rao, and J.~Huang.
\newblock Anti-forensics of camera identification and the triangle test by
  improved fingerprint-copy attack.
\newblock {\em arXiv preprint arXiv:1707.07795}, 2017.

\bibitem{logan2009automatic}
R.~K. Logan, R.~S. Szeliski, and M.~T. Uyttendaele.
\newblock Automatic digital image grouping using criteria based on image
  metadata and spatial information, August 2009.
\newblock US Patent 7,580,952.

\bibitem{Lowe_1999}
D.~Lowe.
\newblock Object recognition from local scale-invariant features.
\newblock In {\em IEEE Conference on Computer Vision and Pattern Recognition},
  volume~2, pages 1150--1157, 1999.

\bibitem{Lowe:IJCV:2004}
D.~G. Lowe.
\newblock Distinctive image features from scale-invariant keypoints.
\newblock {\em International Journal of Computer Vision}, 60(2):91--110, 2004.

\bibitem{mahdian2010bibliography}
B.~Mahdian and S.~Saic.
\newblock A bibliography on blind methods for identifying image forgery.
\newblock {\em Elsevier Signal Processing: Image Communication},
  25(6):389--399, 2010.

\bibitem{marra2014attacking}
F.~Marra, F.~Roli, D.~Cozzolino, C.~Sansone, and L.~Verdoliva.
\newblock Attacking the triangle test in sensor-based camera identification.
\newblock In {\em IEEE International Conference on Image Processing}, pages
  5307--5311, 2014.

\bibitem{Matas_2004}
J.~Matas, O.~Chum, M.~Urban, and T.~Pajdla.
\newblock {Robust Wide Baseline Stereo from Maximally Stable Extremal Regions}.
\newblock {\em Elsevier Image and Vision Computing}, 22(10):761--767, 2004.

\bibitem{monetbords}
A.~Mohdin.
\newblock {Is this Monet real or fake-—and who gets to decide?}
\newblock
  \url{https://qz.com/588932/is-this-monet-real-or-fake-and-who-gets-to-decide/},
  2016.
\newblock Accessed on 06-25-2018.

\bibitem{moreira2018image}
D.~Moreira, A.~Bharati, J.~Brogan, A.~Pinto, M.~Parowski, K.~W. Bowyer, P.~J.
  Flynn, A.~Rocha, and W.~J. Scheirer.
\newblock Image provenance analysis at scale.
\newblock {\em arXiv preprint arXiv:1801.06510}, 2018.

\bibitem{mediscore}
{National Institute of Standards and Technology}.
\newblock {MediScore:Scoring tools for Media Forensics Evaluations}.
\newblock \url{https://github.com/usnistgov/MediScore}, 2017.
\newblock Accessed on 07-01-2018.

\bibitem{nist2017dataset}
{National Institute of Standards and Technology}.
\newblock {Nimble Challenge 2017 Evaluation}.
\newblock
  \url{https://www.nist.gov/itl/iad/mig/nimble-challenge-2017-evaluation},
  2017.
\newblock Accessed on 10-05-2017.

\bibitem{nist2017plan}
{National Institute of Standards and Technology}.
\newblock {Nimble Challenge 2017 Evaluation Plan}.
\newblock
  \url{https://www.nist.gov/sites/default/files/documents/2017/09/07/\\nc2017evaluationplan\_20170804.pdf},
  2017.
\newblock Accessed on 06-18-2018.

\bibitem{oliveira2014multiple}
A.~Oliveira, P.~Ferrara, A.~De~Rosa, A.~Piva, M.~Barni, S.~Goldenstein,
  Z.~Dias, and A.~Rocha.
\newblock Multiple parenting identification in image phylogeny.
\newblock In {\em IEEE International Conference on Image Processing}, pages
  5347--5351, 2014.

\bibitem{pan2004automatic}
J.-Y. Pan, H.-J. Yang, P.~Duygulu, and C.~Faloutsos.
\newblock Automatic image captioning.
\newblock In {\em IEEE International Conference on Multimedia and Expo},
  volume~3, pages 1987--1990, 2004.

\bibitem{pintoFiltering}
A.~Pinto, D.~Moreira, A.~Bharati, J.~Brogan, K.~Bowyer, P.~Flynn, W.~Scheirer,
  and A.~Rocha.
\newblock Provenance filtering for multimedia phylogeny.
\newblock In {\em IEEE International Conference on Image Processing}, pages
  1502--1506, 2017.

\bibitem{fakenewsmexico}
M.~Pskowski.
\newblock Mexico struggles to weed out fake news ahead of its biggest election
  ever.
\newblock
  \url{https://www.theverge.com/2018/6/27/17503444/mexico-election-fake-news-facebook-twitter-whatsapp},
  2018.
\newblock Accessed on 06-28-2018.

\bibitem{reddit2017photoshopbattles}
{Reddit.com}.
\newblock {Photoshopbattles}.
\newblock \url{https://www.reddit.com/r/photoshopbattles/}, 2017.
\newblock Accessed on 08-11-2017.

\bibitem{sasikala2015efficient}
S.~Sasikala and R.~S. Gandhi.
\newblock Efficient content based image retrieval system with metadata
  processing.
\newblock {\em International Journal of Innovative Research in Science and
  Technology}, 1(10):72--77, 2015.

\bibitem{simmhan2005survey}
Y.~L. Simmhan, B.~Plale, and D.~Gannon.
\newblock A survey of data provenance techniques.
\newblock \url{ftp://ftp.extreme.indiana.edu/pub/techreports/TR618.pdf}, 2005.
\newblock Accessed on 06-13-2018.

\bibitem{provimpnews}
R.~D. Spencer and G.~D. Sesser.
\newblock {Provenance: Important, Yes, But Often Incomplete and Often Enough,
  Wrong}.
\newblock
  \url{https://news.artnet.com/market/the-importance-of-provenance-in-determining-authenticity-29953},
  2013.
\newblock Accessed on 06-22-2018.

\bibitem{metprovproject}
{The Metropolitan Museum of Art}.
\newblock {The Met's Provenance Research Project}.
\newblock
  \url{https://www.metmuseum.org/about-the-met/policies-and-documents/provenance-research-project},
  200.
\newblock Accessed on 06-22-2018.

\bibitem{tsai2005extent}
C.-M. Tsai, A.~Qamra, E.~Y. Chang, and Y.-F. Wang.
\newblock Extent: Inferring image metadata from context and content.
\newblock In {\em IEEE International Conference on Multimedia and Expo}, pages
  1270--1273, 2005.

\bibitem{wikieiffel}
{Wikimedia Commons}.
\newblock {Las Vegas' Eiffel Tower}.
\newblock
  \url{https://commons.wikimedia.org/wiki/File:Eiffel_Tower_in_Las_Vegas_(Paris)_at_night.jpg},
  2015.
\newblock Accessed on 06-25-2018.

\bibitem{yee2003faceted}
K.-P. Yee, K.~Swearingen, K.~Li, and M.~Hearst.
\newblock Faceted metadata for image search and browsing.
\newblock In {\em ACM SIGCHI Conference on Human Factors in Computing Systems},
  pages 401--408, 2003.

\bibitem{zagoruyko_2015}
S.~Zagoruyko and N.~Komodakis.
\newblock Learning to compare image patches via convolutional neural networks.
\newblock In {\em IEEE Conference on Computer Vision and Pattern Recognition},
  pages 4353--4361, 2015.

\bibitem{zhang2018shufflenet}
X.~Zhang, X.~Zhou, M.~Lin, and J.~Sun.
\newblock Shufflenet: An extremely efficient convolutional neural network for
  mobile devices.
\newblock In {\em IEEE Conference on Computer Vision and Pattern Recognition},
  2018.

\end{thebibliography}
}

\end{document}